\documentclass[conference]{IEEEtran}
\IEEEoverridecommandlockouts
\usepackage{cite}
\usepackage{amsmath,amssymb,amsfonts}
\usepackage{algorithmic}
\usepackage{graphicx}
\usepackage{textcomp}
\usepackage{xcolor}
\def\BibTeX{{\rm B\kern-.05em{\sc i\kern-.025em b}\kern-.08em
    T\kern-.1667em\lower.7ex\hbox{E}\kern-.125emX}}
\begin{document}

\title{Preference Change in Persuasive Robotics}

\author{\IEEEauthorblockN{1\textsuperscript{st} Matija Franklin}
\IEEEauthorblockA{\textit{University College London}\\
matija.franklin@ucl.ac.uk}
\and
\IEEEauthorblockN{2\textsuperscript{nd} Hal Ashton}
\IEEEauthorblockA{\textit{University College London}}
}

\maketitle

\begin{abstract}
Human-robot interaction exerts influence towards the human, which often changes behavior. This article explores an externality of this changed behavior - preference change. It expands on previous work on preference change in AI systems. Specifically, this article will explore how a robot's adaptive behavior, personalized to the user, can exert influence through social interactions, that in turn change a user's preference. It argues that the risk of this is high given a robot's unique ability to influence behavior compared to other \textit{pervasive technologies}. \textit{Persuasive Robotics} thus runs the risk of being manipulative.
\end{abstract}


\section{Introduction}
Modern social robotics uses machine learning (ML) methods to learn user preferences in order to develop adaptive robot behavior which is tailored to the user. During human-robot interaction (HRI), robots can learn human preferences by inferring them through observing human behavior in various contexts and tasks \cite{b1}. This approach of learning preferences through inference from behavior is known as \textit{Revealed Preference Theory}. Robots can also learn preferences by asking users directly (e.g., providing a ranking) \cite{b2}. This is known as learning from \textit{Stated Preferences}.

This article argues that any attempt to adapt a robot's behavior to human preference needs to acknowledge that the robot can change human preference. This is due to the fact that although preference does influence behavior, behavior can predate and lead to the formation of new preference \cite{b3}. Certain forms of HRI thus run the risk of being manipulative if the Robot has some preference over human behaviour. It is not possible to ensure that HRI is transparent, ethical and safe without understanding the impact it has on preference. This article will review HRI's influences on behavior, and concentrate on the problem of preference change.

\section{Uniqueness of Robot Influence}

A key difference with robot influence separating it from other forms of \textit{pervasive technology} is a robot's physical embodiment; triggering aspects of human social cognition that are attuned to social influence \cite{b4}. The robot's physical embodiment also allows it to collect rich interaction data that can be used to infer human intention and emotion \cite{b5}. \textit{Persuasive Robotics} studies influence in HRI, specifically focusing on aspects of social interaction (both human-to-human, and human-to-robot) that significantly alter a robot's influence \cite{b6}. 

Compared to other pervasive technologies, such as recommender systems or smart user interfaces, robots additionally influence through social interaction and social presence. Evidence suggests that people form different relationships with robots than they do to virtual avatars and computers. For example, people rate physical robots as more watchful and enjoyable \cite{b7}. People also empathize more with an embodied robot than a virtual robot when watching the robot experience pain \cite{b8}. Finally, a robot's physical embodiment can produce arousing physiological reactions in users \cite{b9}.

The particular relationship people have with robots, compared to other technologies, results in a greater behavior change. There is evidence that people are more likely to follow instructions from a robot than from a computer tablet due to a greater desire to interact with the robot \cite{b10}. Another study replicated this, finding that a greater preference towards a robot, compared to a computer, leads participants to interact with it for longer \cite{b11}.

Sociocognitive factors that influence behaviour in human-to-human interaction, such as inter-group, intra-group and interpersonal factors, are also prominent in HRI. For example, most people after initially interacting with a humanoid social robot will perceive a greater social presence from it \cite{b12}. As with influence exerted by human groups, people will conform to a group of robots by changing their preliminary answers to match the robots' answers \cite{b13}. Further, people will show more positive reactions towards an \textit{in-group} robot versus an \textit{out-group} robot as they will anthropomorphize it more \cite{b14}.

Interpersonal factors and affect also impact human behavior in HRI. Interpersonal, robot-delivered interactions can be as effective as those delivered by humans \cite{b15}. People tend to rate a robot of the opposite sex as more trustworthy, credible, and engaging, with male participants being more likely to donate money to a female robot \cite{b6}. Further, touch, perceived autonomy, and interpersonal distance all have an impact on human behavior \cite{b16}. Finally, robots can influence human behavior with \textit{affective displays} (e.g., conveying distress) \cite{b17}.

\section{Adaptive Robots Change Human Preference}

Social robot's adaptive behavior, tailored to a user's preference, changes the human user's behavior \cite{b18}. Behavioral Science researches \textit{behavioral insights} - cause and effect understanding of how different factors influence behavior - which has allowed it to build valid and reliable predictive models of behavior \cite{b19}. In human-robot interaction, behavioral insights are adjacent to \textit{human factors}. Human-robot interaction with ML-powered systems whose design has been centered around human factors leads to consistent, predictable behavior change.

The practice of learning a user's preference and adapting a social robot's behavior, which in turn changes the behavior of the user, also changes the user's preference. To understand why this is the case, it is important to note that preferences are not static, but rather quite changeable, and predicatively influenced by various factors \cite{b20, b21}. To give an example, one person's preference can change between contexts due to pressure exerted by the social norms of their 'in-group' \cite{b22}. The fact that a person can have multiple preferences in different contexts raises questions related to which one should be thought of as the 'true' preference \cite{b23}. 

It is also important to note that although preference does influence behavior, behavior can predate and lead to the formation of new preference \cite{b3}. Adapting a social robot's behavior to a user preference is not only a matter of preference learning; because the adapted robot behavior changes a user's behavior, it also can and will change a user's preferences. Previous work has explored the problem of behavior and preference manipulation in AI systems; specifically, how iterative ML systems tasked with learning user preferences over time, often impact the preferences they are changing, or worse manipulate them to serve their own objective function \cite{b24, b25}. We thus propose that a multidisciplinary endeavor should research how preference changes - Preference Science \cite{b20}. This includes factoring in the confounding factors that influence both preference and behavior. Future paradigms in HRI should explore the factors that can be highly manipulative over a user's preference.

\section{Conclusion}

This article aimed to outline an ethical issue pertaining to robot manipulation. Specifically, the embodied nature of robots makes HRI additionally influential compared to interaction with other pervasive technologies. Any change in behaviour induced by a robot results in the formation of new preferences. Robots that learn user preferences are thus likely to impact them. They can also manipulate preferences to suit their own objective function, by making people more predictable so as to more easily anticipate their wants and needs. Preference learning thus poses many challenges for developers aiming to design ethical systems for persuasive robotics.

\end{document}